\documentclass[runningheads]{llncs}

\usepackage{eccv}

\usepackage{eccvabbrv}

\usepackage{graphicx}
\usepackage{booktabs}
\usepackage{multirow}
\usepackage{bbm}
\usepackage{subfiles}
\usepackage[accsupp]{axessibility}
\usepackage{makecell}
\usepackage{hyperref}

\usepackage{orcidlink}

\begin{document}

\title{Learning from the Web: Language Drives Weakly-Supervised Incremental Learning for Semantic Segmentation}

\titlerunning{Learning from the Web: Language Drives Weakly-Supervised ILSS}

\author{Chang Liu\inst{1}\orcidlink{0000-0002-5321-2264} \and
Giulia Rizzoli\inst{2}\orcidlink{0000-0002-1390-8419} \and
Pietro Zanuttigh\inst{2}\orcidlink{0000-0002-9502-2389} \and
Fu Li\inst{1}\orcidlink{0000-0003-0319-0308} \and
Yi Niu\inst{1}\orcidlink{0000-0002-7359-276X}\thanks{Corresponding author}}

\authorrunning{C.~Liu, G.~Rizzoli et al.}

\institute{School of Artificial Intelligence, Xidian University, China \and
Department of Information Engineering, University of Padova, Italy
}

\maketitle

\begin{abstract}

Current weakly-supervised incremental learning for semantic segmentation (WILSS) approaches only consider replacing pixel-level annotations with image-level labels, while the training images are still from well-designed datasets. In this work, we argue that widely available web images can also be considered for the learning of new classes. To achieve this, firstly we introduce a strategy to select web images which are similar to previously seen examples in the latent space using a Fourier-based domain discriminator.  Then, an effective caption-driven reharsal strategy is proposed to preserve previously learnt classes. To our knowledge, this is the first work to rely solely on web images for both the learning of new concepts and the preservation of the already learned ones in WILSS. Experimental results show that the proposed approach can reach state-of-the-art performances without using manually selected and annotated data in the incremental steps. 

\end{abstract}

\begin{figure}[t]
\centering
\includegraphics[trim=1cm 6.5cm 4.5cm 6cm, clip, width=\linewidth]{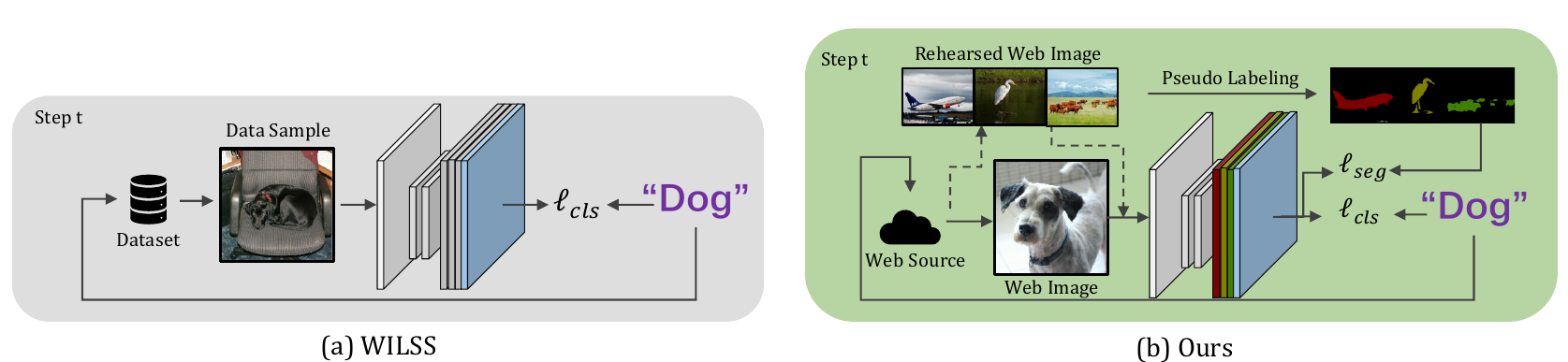}
\caption{General overview of the proposed method.} 
\label{fig:PvsWP}
\end{figure}

\section{Introduction}
\label{sec:intro}
The rapid development of deep learning techniques allowed for impressive results in Semantic segmentation (SS), i.e., pixel-level labeling of real-world scenes.
Traditional supervised SS approaches are constrained to predetermined classes, leading to the problem of \textit{catastrophic forgetting} when learning from new data. To address this, incremental learning techniques, such as Class Incremental Learning for Semantic Segmentation (CILSS), have been proposed. Several CILSS strategies aim to balance new and old knowledge by using regularization strategies or storing past examples \cite{michieli2019incremental,cermelli2020modeling}. However, these methods heavily rely on expensive pixel-level annotations.
Weakly-Supervised Incremental Learning for Semantic Segmentation (WILSS) has recently emerged to reduce annotation costs, utilizing image-level labels for guidance. Nevertheless, existing WILSS approaches face limitations, such as requiring multiple classes at each step and performance degradation with multiple steps \cite{cermelli2022incremental,yu2023foundation}.
In this study, we focus on a novel paradigm, Web-based Weakly-Supervised Incremental Learning, which leverages both existing knowledge and new information from the web. This approach finds applications across various domains, particularly when pre-trained models need adaptation to new classes where data are limited or unavailable, e.g., due to privacy concerns \cite{shenaj2023federated}.
Leveraging web resources becomes crucial to supplement the existing model with additional class information, enabling adaptation without extensive data collection efforts.
However, within this framework, new challenges emerge:
1) Although external sources such as the internet provide a potential solution, web data often come with noisy image-level supervision, containing undesired classes or unlabeled classes that were previously learned.  We tackle this issue by selecting images using adversarial learning in the Fourier domain and leveraging language models to provide multi-class label supervision.
2) Additionally, without access to previous training data, the model may suffer from catastrophic forgetting. 
To address this issue, we propose leveraging web resources also for preserving the proficiency of the model on old classes.  Nevertheless, integrating this data into the existing model poses challenges, especially when the new data deviates from the original distribution. This misalignment affects the training process, requiring innovative strategies to ensure effective model adaptation to the new classes. To this extent, instead of querying through the image label (i.e., the class name), we used captions that better represent the image content. 
An overview of our framework is shown in \cref{fig:PvsWP}, in summary the main contributions are:
\begin{itemize}
    \item We present a comprehensive framework that leverages web images to tackle the challenges of learning new classes in an incremental fashion and of knowledge preservation in scenarios with limited or inaccessible training data;
    \item By integrating vision-language captioning, we enhance image-level supervision with textual descriptions;
    \item By downloading images using vision-language captioning, we can improve the searched queries and re-generate them to further  verify that images have the expected content.
    \item Concluding, we introduce a new flexible data pipeline that allows to train continual learning schemes even without resorting to curated datasets.
\end{itemize}

\section{Related Work}
\label{sec:relat}
\textbf{Class-Incremental Learning for SS (CILSS)} 
Class-incremental continual learning has been extensively explored within the image classification domain, with recent efforts extending into the more complex task of semantic segmentation \cite{michieli2019incremental}. Initially, techniques such as knowledge distillation, parameters freezing, and class re-weighting were employed to introduce new classes while retaining knowledge from prior ones \cite{michieli2019incremental,klingner2020class,cermelli2020modeling}.
Subsequently, researchers delved into regularization and contrastive mechanisms at the feature level to enhance class-specific capabilities and maintain spatial relationships \cite{michieli2021continual,douillard2021plop,yang2022uncertainty}. Further advancements include the proposal of novel objective functions to account for class similarity \cite{phan2022class}, and biased-context-insensitive consistency \cite{zhao2022rbc}.
Other proposed strategies consider dynamic architectures to balance the network parameters \cite{zhang2022representation,xiao2023endpoints} and rehearsal data to preserve old knowledge \cite{liu2023recall+}.\\
\noindent \textbf{Weakly Supervised Incremental Learning for SS (WILSS)} Weakly Supervised Semantic Segmentation techniques aim to develop effective segmentation models using more affordable forms of supervision, such as bounding boxes \cite{dai2015boxsup,song2019box}, scribbles \cite{lin2016scribblesup}, saliency maps \cite{wang2020self,lee2021railroad}, points \cite{mcever2020pcams}, and image-level labels \cite{ahn2018learning,araslanov2020single}. Following this concept, Weakly Supervised Incremental Learning for Semantic Segmentation (WILSS) involves enhancing pre-trained segmentation models with new classes using only image-level supervision. To this extent, WILSON \cite{cermelli2022incremental} proposes a comprehensive framework, which can learn progressively from pseudo-labels generated online by a localizer attached to the model. FMWISS \cite{yu2023foundation} seeks to enhance pseudo-labels by leveraging complementary foundational models. Despite the efforts, both methods are unable to perform single-class incremental learning steps, requiring negative examples to properly guide the training.
Lastly, RaSP \cite{roy2023rasp} relies on foundation models to build semantic connections between image-level labels.
\\
\noindent\textbf{Webly-Supervised Learning} is an emerging field in research where deep learning models are trained using large amounts of data from the web \cite{chen2015webly,niu2018webly,sun2023webly}.
Current research efforts are focused on understanding how to effectively query and select samples \cite{duan2020omni,qin2024datasetgrowth} and how to leverage weakly supervised data \cite{luo2020webly,yang2020webly}, e.g., by generating pseudo-labels.
In CILSS, challenges arise due to the absence of pixel-level supervision and the presence of distribution shifts compared to the original data. The first approach to exploit web data in CILSS is RECALL \cite{maracani2021recall,liu2023recall+}, which investigates preserving previous knowledge by leveraging pixel-level pseudo-labels obtained through either web data querying or the generation of synthetic samples.
However, it is essential to acknowledge that utilizing queried images not only lacks pixel-level labeling but also introduces inconsistency at the image-level. In our work, we specifically address the scenario where images may contain multiple classes, despite being queried and labeled under a single class label. This aspect holds particular significance in WILSS - as new knowledge is solely acquired from image-level labels - and incorrect supervision significantly worsens the results.
Moreover, to our knowledge, this is the first work exploring the usage of web samples for both learning new knowledge and preserving previous one. \\ 
\noindent \textbf{Foundation Models in WILSS}
The early stage of pre-trained foundation models starts from the text field, such as BERT \cite{devlin2018bert}, before expanding into the domain of visual language models (VLMs) \cite{pmlr-v139-radford21a}.
Recent years have witnessed a notable shift towards the integration of image and text, leading to active research in VLMs for a variety of tasks, including text-to-image generation \cite{li2023gligen}, zero-shot segmentation \cite{zhou2016learning,kirillov2023segment}, and visual question answering \cite{ravi2023vlc}.
CLIP \cite{pmlr-v139-radford21a} serves as a milestone in VLM development, where both image and text encoders are jointly trained to align image-text pairs. Motivated by this, recent studies have integrated VLMs into frameworks like WILSS \cite{yu2023foundation,roy2023rasp}. Yu et al. \cite{yu2023foundation} exploit the zero-shot capabilities of foundational models, combining DINO \cite{caron2021emerging} and MaskCLIP \cite{zhou2022extract} for pseudo-supervision. Instead, Roy et al. utilize BERT \cite{devlin2018bert} to derive textual descriptions and compute semantic matches between new and old classes.
In contrast, our approach utilizes a captioning model, specifically Flamingo \cite{Alayrac2022FlamingoAV, awadalla2023openflamingo}, to extract meaningful captions solely from image inputs. Given its capability to adapt to novel tasks, we leverage it for tasks such as providing multi-label supervision, querying, and refining web data.
\section{Method}
\label{sec:method}

\begin{figure}[ht]
\centering
\includegraphics[trim=4.5cm 3.3cm 6cm 1.8cm, clip, width=.85\linewidth]{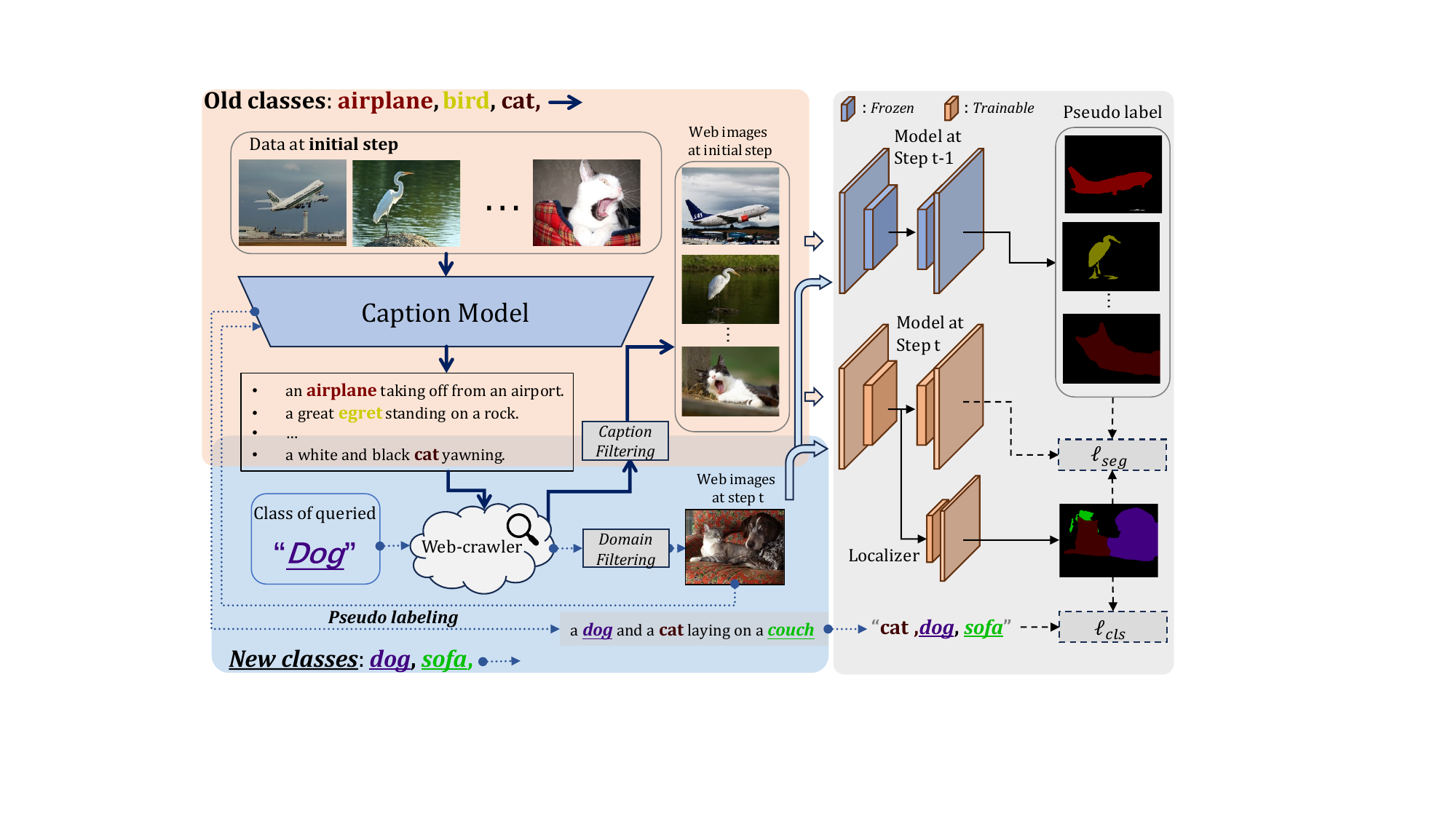}
\caption{The proposed method employs a web crawler to gather new knowledge and retrieve past data.  The new knowledge (\cref{sec:learn_new}) is acquired by querying class names and subsequently filtered in the Fourier domain. Simultaneously, image-level labels are provided by a captioning model. On the other hand, the preservation of old knowledge (\cref{sec:learn_old}) involves querying captions from previous data and filtering them based on semantic similarity with the regenerated captions.} 
\label{fig:method}
\vspace{-0.35cm}
\end{figure}

In the considered setting we assume to have access to images in a fully-supervised dataset only for the first training phase on an initial set of classes, after which it is no longer available. Instead, data for newly identified classes are exclusively sourced from external platforms, notably the Internet\footnote{Images were downloaded from the \textit{Flickr} website adhering to its \href{https://www.flickr.com/help/terms}{term of use}.}. Underlying this setup is the assumption that only the names of the new target classes are known, without any supplementary guidance. In Section \ref{sec:pd}, we outline the problem and the considered framework, as established in \cite{cermelli2022incremental}. Subsequently, in Sections \ref{sec:learn_new} and \ref{sec:learn_old} we present the novel proposed scenario and our approach to tackling it by encompassing previous limitations. An outline of our method is in \cref{fig:method}.

\subsection{Problem Definition}
\label{sec:pd}
\textbf{WILSS task.} Lets consider an image $\textbf{x} \in \mathcal{X} \subset \mathbb{R}^{|\mathcal{I}| \times 3}$ drawn from a distribution $\mathcal{D}$, where $|\mathcal{I}| = H \times W$. Its semantic description can be in the form of an image-level label $y \in \mathcal{Y} \subset \mathcal{C}$ (we represent it using one-hot vectors, i.e., $y \in \mathbb{R}^{C}$) or of a more detailed pixel-level segmentation map $\textbf{y} \in \mathbb{R}^{|\mathcal{I}| \times C}$. The label set $\mathcal{C} = \{b, c_1, \dots, c_{C-1}\}$ is the same for image-level and pixel-level labeling and contains $C$ target classes ($b$ is the background class).

In the class-incremental setting, the training is assumed to happen over a sequence of steps $t = 0, \dots, T$. 
In contrast to pixel-level segmentation, within the framework of WILSS, at the initial step ($t=0$) pixel-level labels $\textbf{y}$ are available, whereas for all subsequent future steps, it becomes solely the image-level label $y \in \mathbb{R}^{C}$.
More in detail, at each step $t>0$, new classes $\mathcal{C}^t$ with image-level annotations are introduced, leading to the creation of a new label set $\mathcal{Y}^t = \mathcal{Y}^{t-1} \cup \mathcal{C}^t$.
As in exemplar-free settings, by the time step $t$ is reached, the images from the previous step ($t-1$) are no longer accessible.

\noindent \textbf{WILSON framework \cite{cermelli2022incremental}.}
At each step $t$, the network is composed of a shared encoder $E^t$, a segmentation decoder $D^t$, which is incrementally extended to accommodate new classes, and a localizer head $L^t$, that is trained from scratch for every task. Additionally, the model from the preceding task is retained and denoted as $(E \circ D)^{t-1}$. 
The localizer is utilized to generate pseudo-supervision for the segmentation model, providing two guidance: one at the image level and the other at the pixel level.  
Let $\textbf{y}_{L}^t \in \mathbb{R}^{|\mathcal{I}| \times |\mathcal{Y}^t|}$  be the pixel-level prediction of the localizer $(E \circ L)^t$. As the localizer is trained from scratch at each step, we simplify the notation by making $\textbf{y}_{L}^t = \mathbf{y}_L$. Additionally, we define $y_{L} \in \mathbb{R}^{|\mathcal{Y}^t|}$ as the image-level counterpart, which is derived from the localizer's scores by using normalized Global Weighted Pooling as in \cite{araslanov2020single}. 
Specifically, the image-level supervision is computed as a multi-label soft-margin loss:
\begin{equation}
    \mathcal{L}_{CLS}(y, y_L) = -\frac{1}{|\mathcal{C}^t|} \sum_{c \in \mathcal{C}^t} y^c log(y_L^c) + (1-y^c) log (1 - y_L^c)
    \label{eq:cls}
\end{equation}
Given that pixel-level predictions from a classifier may contain noise, $\textbf{y}_{L}$ is initially smoothed \cite{lukasik2020does} ($\tilde{\textbf{y}}_L$) - please refer to \cite{cermelli2022incremental} - and then combined with the old model predictions:
\begin{equation}
    \hat{\textbf{y}} = 
    \begin{cases}
        \min(\textbf{y}_D^{c(t-1)}, \tilde{\textbf{y}}^c_L) & \text{if } c = b\\
        \tilde{\textbf{y}}^c_L & \text{if } c \in \mathcal{C}^t \\
        \textbf{y}_D^{c(t-1)} & \text{otherwise}
    \end{cases}
\end{equation}
where $y_{D}^{t-1} \in \mathbb{R}^{|\mathcal{I}| \times |\mathcal{Y}^{t-1}|}$ is the prediction of the segmentation model $(E \circ D)^{t-1}$.
Then, the pixel-level loss is obtained as:
\begin{equation} \label{eq:seg}
    \mathcal{L}_{SEG}(\hat{\mathbf{y}}, \mathbf{y}_D^t) = - \frac{1}{|\mathcal{I}| |\mathcal{Y}^t|} \sum_{i \in \mathcal{I}} \sum_{c \in \mathcal{Y}^t} \hat{{y}}^c_{(i)} log({y}_{(i)D}^{c \, t}) + (1-\hat{{y}}^c_{(i)}) log (1 - {y}_{(i)D}^{c \, t})
\end{equation}
where we denoted with ${\hat{y}_{(i)}}$ the predictions for the i-th pixel.
Furthermore, to prevent the forgetting of old classes,  two additional knowledge distillation losses are proposed in \cite{cermelli2022incremental}: 1) $\mathcal{L}_{KDE}$ aligns the features from the current encoder $E^t$ with the previous one $E^{t-1}$ through mean-squared error; 2) $\mathcal{L}_{KDL}$ improves consistency between the old model's $(E \circ D)^{t-1}$  and the localizer's $(E \circ L)^{t}$.
Ultimately the objective function is computed as follows: $\mathcal{L} = \mathcal{L}_{SEG} + \mathcal{L}_{CLS} + \mathcal{L}_{KDE} + \mathcal{L}_{KDL}$ (see the supplementary material for details).

\subsection{Learning new Knowledge from Web}
\label{sec:learn_new}
In CILSS, the web has been used for rehearsal of past classes \cite{maracani2021recall,liu2023recall+}. The samples are gathered from the web by querying each class using the corresponding class name, obtaining the corresponding image set:
\begin{equation} \label{eq:download}
    \mathcal{X}^{web} = \{ \textbf{x} = \mathcal{D}^{web}(q) \, | \, q = c \}
\end{equation}
where $q$ is the searched query. 
As the query $q$ corresponds to the class-name of $c$, the naive way to provide image-level supervision for $\textbf{x} \in \mathcal{X}^{web}$ is:
\begin{equation} \label{eq:naivelabeling}
    y^c = \begin{cases}
    1 & \text{if } q = c \\
    0 & \text{otherwise}
    \end{cases}
\end{equation}
We introduce a scenario wherein both the images from the preceding step \( t - 1 \) and those from the current step \( t \) are unavailable (i.e., access to the original dataset distribution \( \mathcal{D} \) is not allowed after the initial step). At each incremental step \( t \), the newly introduced image \( \mathbf{x} \in \mathcal{X}^{web} \subset \mathbb{R}^{|\mathcal{I}| \times 3} \) belonging to the new class set \( \mathcal{C}^t \) is sampled from a distribution \( \mathcal{D}^{web} \) distinct from \( \mathcal{D} \), thus representing an external data source. Notably, in our framework, the test dataset belongs to the original distribution $\mathcal{D}$.
Real dataset images are typically manually selected and verified, resulting in superior content quality and greater diversity. Furthermore, in most cases, their labeling is the product of costly human labour.
To explore learning effectively from web images, firstly a Fourier transformation-based domain discriminator is designed for selecting web images which are considered to have more similarities with previous samples in latent space. 
Then, a pre-trained image caption model is adopted for generating pseudo image-level annotations.

\subsubsection{Fourier Domain-based Discriminator}
Although web images are widely available, by downloading a small sample from the internet there is no guarantee that its distribution resembles the one from the dataset with which we pretrained the model at step $0$. 
In order to preserve the same statistics from the original dataset, we propose to train a discriminator network to distinguish between the two distributions, selecting web images that closely resemble those from the original dataset. 
However, since no dataset images are available for the current classes in the incremental steps, the discriminator must be trained solely in the initial step and continue to function effectively in subsequent steps. 

Based on the above concern, we introduce a Fourier-based domain discriminator.
The rationale behind this approach is that amplitude and phase representations in the Fourier domain tend to exhibit more consistency in class statistics changes compared to the pixel domain. Recent domain adaptation techniques have leveraged this consistency, applying amplitude as a stylization to preserve the style of the target distribution \cite{yang2020fda,rizzoli2024source}.
Given an input $\mathbf{x}$, the prediction from discriminator is:
\begin{equation}
    (p_{ds}, \, p_{web}) = (p(\mathbf{x} \in \mathcal{X}), \, p(\mathbf{x} \in \mathcal{X}^{web}))=M_D(| \mathcal{F}(\mathbf{x})|)
\end{equation}
where the two outputs correspond to the scores for the data to belong to the source dataset (e.g., PASCAL-VOC in our experiments) or to the web, respectively. $\mathcal{F}$ represents the Fourier transform operator, the modulus operation extracts the amplitude value of the transformed sample, and $M_D$ is the proposed discriminator. In this context, the corresponding amplitude value of the transformed sample is fed into the discriminator. An image is considered valid if it can deceive the discriminator, meaning $p_{ds}/p_{web} > 1$.

\subsubsection{Caption Labeling} \label{sec:learn_new:labeling}

At each step $t$, the localizer has to identify objects within the image. 
Crucially, if the localizer predicts a new class for a given pixel, it is assigned as such, thus ensuring adaptability to novel classes.  
One of the key challenges we address is the reliance on image-level labels for supervising the localizer. When we introduce unsupervised images for the web instead of the data from the original distribution, as images are downloaded querying the class name and labelled with it (as in \cref{eq:naivelabeling}), 
these labels are prone to inaccuracies, leading to potential misclassifications.
Inspired by the remarkable performance of pre-trained vision language models (VLMs), we introduce a strategy that leverages a VLM to enhance image-level supervision.
Specifically, we utilize a captioning model \( M_{CAP}\), which takes as input the image \( \mathbf{x} \in \mathcal{X} \subset \mathbb{R}^{|\mathcal{I}| \times 3} \) drawn from a distribution \( \mathcal{D}^{web} \), where \( \mathcal{D}^{web} \cap \mathcal{D} = \varnothing \).
First, we consider a set of words $\mathcal{W}^c$ for each class $c \in \mathcal{Y}^t$, including synonyms, plurals and linguistic alternatives 
(e.g. ``plane'' belongs to the ``aeroplane'' class, ``woman'' belongs to the ``person'' class, refer to the supplementary material for the full list).
Then, we generate the caption corresponding to each image: $w = M_{CAP}(\textbf{x})$.
The image-level label $ y \in \mathbb{R}^{C}$  is derived from \( w \), $ \forall \, c \in \mathcal{Y}^t$, as:
\begin{equation}
    y^c = \begin{cases}
    1 & \text{if } \exists w_i \in w : w_i \in \mathcal{W}^c \\
    0 & \text{otherwise}
\end{cases}
\end{equation}
providing multi-label guidance to the web images.
For example, a caption mentioning ``a person standing on a boat'' would yield labels for both ``person'' and ``boat''. This approach not only enhances the granularity of image-level supervision but also addresses inconsistencies in labeling. Moreover, it allows us to identify and discard images that lack the relevant class altogether, thus improving the overall quality of the dataset.

\subsection{Rehearsal strategies}
\label{sec:learn_old}
Several existing WILSS methodologies \cite{cermelli2022incremental,yu2023foundation} face significant limitations: 1) they require the availability of multiple classes at each incremental step. 
2) their performance drops largely when multiple steps are performed. 
To mitigate these issues, we propose a rehearsal strategy that avoids storing original images, which may raise privacy concerns. 
Specifically, we explore the feasibility of saving solely the image captions and subsequently querying relevant samples from the web. 
Furthermore, we utilize the caption model for additional supervision to verify whether the downloaded images contain the primary class for which they were obtained. This approach not only addresses potential privacy concerns associated with storing original images but also provides a practical solution for overcoming the constraints observed in conventional WILSS methods.
Notably, to adapt WILSS to training on old class data, we modify \cref{eq:cls} and \cref{eq:seg}. Specifically, we define $\textbf{x} \in \mathcal{X}_{r}$ where $\mathcal{X}_{r}$ is the set of web rehearsal data. As the image $\textbf{x}$ should contain  previous classes, the segmentation loss is simply $\mathcal{L}_{SEG}(\textbf{y}_D^{t-1}, \textbf{y}_D^t)$, while the classification loss of the localizer is  $\mathcal{L}_{CLS}(y_D^{t-1}, y_L)$, where the image level label $y_D^{t-1}$ comes from the predicted classes in the pixel-level label $\textbf{y}_D^{t-1}$ (i.e., $y_D^{c \, (t-1)} = 1$ if class $c$ is present in $\textbf{y}_D^{t-1}$).

Automatically obtaining web images suitable for training the network is challenging due to the lack of control on the retrieval process. 
Having previously utilized the image-level capabilities of a VLM to enhance image-level supervision, we decided to exploit its ability to generate representative text descriptions for previously trained samples. 
This brings us three main advantages: 1) we can store the text-level caption of the image sample instead of the image itself, thus strongly reducing storage and privacy concerns; 2) web images queried with accurate captions contain more classes and share more similar semantic content with respect to dataset samples if compared with image queried with just the class name;  3) by re-generating descriptions for queried images, one can easily discard images that do not contain content related to the considered class. 
The strategy consists  of 2 steps: caption-based querying and caption-based filtering.

\subsubsection{Caption-based Querying}
We employed the same model discussed in \cref{sec:learn_new:labeling} for captioning, denoted as \( M_{CAP} \). In the initial step, we obtained captions for all inputs \( \textbf{x} \) belonging to the set \( \mathcal{X} \) such that \( y \) is in \( \mathcal{Y}^{0} \). Then, the new downloaded images are obtained as:

\begin{equation} \label{eq:downloadcaption}
    \mathcal{X}^{web}_{r} = \{ \textbf{x} = \mathcal{D}^{web}(q') \, | \, q' = M_{CAP}(\textbf{x}) : \textbf{x} \in \mathcal{X} \}
\end{equation} 
replacing the naive downloading process described in \cref{eq:download}. Additionally, leveraging replay data, at step \( t \), we have images from classes \( \mathcal{Y}^{t-1} \). Therefore, concerning \cref{eq:cls}, we train the localizer not only on \( \mathcal{C}^t \) but also on \( \mathcal{Y}^{t} \). 
Specifically, we denote the memory capacity as \( M \), i.e., the number of rehearsal images considered for each class. 

\begin{figure}[t]
\centering
\includegraphics[trim=6cm 6.5cm 9cm 6.3cm, clip, width=0.9\linewidth]{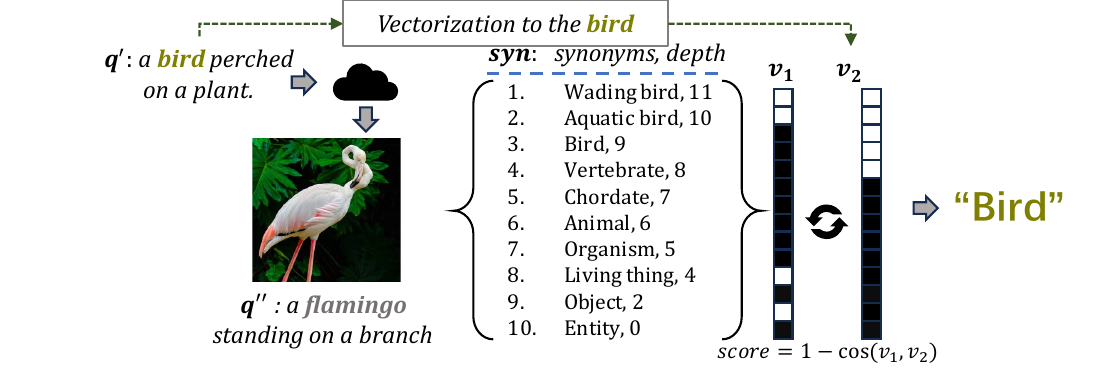}
\caption{Illustration of caption-based filtering approach.} 
\label{fig:cap-filter}
\end{figure} 

\subsubsection{Caption-based Filtering} Despite querying \( \textbf{x} \in \mathcal{X}^{web}_{r} \) based on a more representative description rather than the class name, there is no guarantee that samples include the desired class. Moreover, there is also no guarantee that the captions contain semantic language corresponding to the original class names, therefore we needed to rely on a different strategy with respect to \cref{sec:learn_new:labeling}. Specifically, inspired by classic NLP methods, we used Penn TreeBank \cite{taylor2003penn} and the lexical database WordNet \cite{fellbaum1998wordnet}.
The approach is structured as follows:
\begin{enumerate}
    \item we re-generate a second caption $q''$ by feeding $\textbf{x} \in \mathcal{X}^{web}_{r}$ into the caption model as $q''= M_{CAP}(\textbf{x})$;
    \item we tokenized the two queries and identified two pairs of words $(n'_1, n'_2)$ and $(n''_1, n''_2)$, which represent the first two nouns from sentences $q'$ and $q''$, respectively, based on their syntax tags as in \cite{taylor2003penn};
    \item for each noun $n$ we extract a set of hypernyms, i.e. its cognitive synonyms (see an example in \cref{fig:cap-filter}),  
    from the WordNet Tree \cite{fellbaum1998wordnet};
    \item for each noun $n$ we create a corresponding vectorized descriptor $v$ (e.g., $v'_1$ corresponds to $n'_1$, and so on) by considering the depth $d$ of each of its hypernyms in the tree and updating $v[d]+=1$; 
    \item we compute the cosine distance between each vector (e.g., we compare $v'_1$ with both $v''_1$ and  $v''_2$);
    \item if at least one couple of descriptors is similar (e.g., $v'_1$ with $v''_1$ or $v''_2$), then the image \textbf{x} is kept, meaning that it contains at least one concept that was contained in the original image. We assume two vectors $v_i$ and $v_j$ to be similar if $1 - cos(v_i,v_j)> T$.
\end{enumerate}
 
Essentially, we assessed whether the words in the captions belong to the same semantic family. While this approach does not guarantee that the words correspond to the set of the old classes, this strategy enforces the same image content from the original distribution, which is not limited to the learnt categories.

\section{Experimental Results}
\label{sec:exper}

\textbf{WILSS Dataset and Settings}
Aligning with WILSON \cite{cermelli2022incremental}, we experimented on PASCAL-VOC (VOC) \cite{everingham2010pascal} and COCO \cite{lin2014microsoft} benchmarks. We considered two setups: 1) Single-step multi-class, with one single incremental step with multiple classes (e.g., \textbf{15-5}, learn 15 classes at step 0, then 5 classes together);
2) multi-step single-class, with multiple incremental steps while learning only one class (e.g., \textbf{15-1}, one class for each incremental step).
In the COCO-to-VOC setting, the initial step is computed on COCO, while the incremental steps on VOC, results for this setting please refer to the Sec. \ref{sec:exp}.  
We adopted two standard settings \cite{michieli2019incremental}:   \textit{disjoint} and \textit{overlapped}. In both, only the image-level labels belonging to the current task are provided (i.e., $y \in \mathcal{C}^t$). In disjoint, for each step $t$, images of the current task contain only classes in $\mathcal{Y}^{t}$, while in overlapped, images can contain any class, even future ones.\\
\textbf{Web Dataset}
We utilize the Flickr website as our primary source for web images due to its extensive collection of diverse images.
However, it is important to note that if licensing poses an issue, alternative publicly available sources can be explored. For each class being learned, a set of 10K web images has been gathered using \cref{eq:download}. For the new knowledge, we select 500 samples for training after implementing the strategies detailed in \cref{sec:learn_new}.  For the old class images, we retrieved 20 images per caption ($N_{CAP}$) using the download caption equation (\cref{eq:downloadcaption}). However, to ensure comparability to the competitors \cite{roy2023rasp}, we employed 100 rehearsal images ($M=100$).\\
\textbf{Implementation Details}
Following \cite{cermelli2022incremental}, we use DeeplabV3 \cite{deeplabv3} with ResNet-101 \cite{resnet101} for Pascal VOC and Wide-ResNet-38 \cite{wu2019wider} for COCO, with output stride equal to 16 and 8 respectively.
The model is trained with SGD for 30 epochs at the initial step and 40 epochs for the following incremental steps. For the first 5 epochs, only the localizer is trained. Then the main model is trained with the pseudo annotation combined with localizer output and previous model.
For the discriminator $M_D$, we adopted the lightweight EfficientNet-B0 model  \cite{efficientnet} followed by 3 fully connected layers with dimensions 1000, 256 and 2 to prevent memory burden. The backbone network has been pre-trained on ImageNet.  
Then, it is trained only at the initial step, with dataset images of initial classes as positive samples and the web images queried with related class names as negative samples. The training lasts for 10 epochs, with image size 224 and batch size 24, optimized by SGD. 
For the captioning model $M_{CAP}$, we employed the pre-trained VLM Openflamingo\cite{awadalla2023openflamingo} as it could be applied to various vision-text tasks using a single model and is publicly open-sourced.  For filtering in \cref{sec:learn_old}, we used the NLTK library \cite{bird2009natural}, which is an NLP tool able to tokenize sentences and parse them in the Penn Treebank \cite{taylor2003penn}. The filtering threshold is set to $T=0.6$. One additional aspect to consider is that when dealing with both new and existing data sourced from the web, the weight of the $L_{KDE}$ loss is reduced to 0.5. 
The code is available at \href{https://github.com/dota-109/Web-WILSS}{https://github.com/dota-109/Web-WILSS}.

\subsection{Experimental Comparison}
\label{sec:exp}
\begin{table}[t]
\centering
\caption{Comparison on single-step multi-class settings on PASCAL-VOC.}
\label{tab:merged_methods}
\resizebox{\textwidth}{!}{
\begin{tabular}{ccccccc|ccc|ccc|ccc}
\toprule
\textbf{} & \textbf{} & \textbf{} &  \textbf{} & \multicolumn{6}{c}{\textbf{15-5}} & \multicolumn{6}{c}{\textbf{10-10}} \\ \cmidrule{5-16} 
 &  &  &  & \multicolumn{3}{c}{\textit{Disjoint}} & \multicolumn{3}{c}{\textit{Overlapped}} & \multicolumn{3}{c}{\textit{Disjoint}} & \multicolumn{3}{c}{\textit{Overlapped}} \\ \cmidrule{5-16} 
\textbf{Method} & \textbf{Train} & \textbf{Memory} & \textbf{M} & 1-15 & 16-20 & All & 1-15 & 16-20 & All & 1-10 & 11-20 & All & 1-10 & 11-20 & All \\ 
\midrule
CAM† \cite{zhou2016learning}                   & VOC & - & - &69.3   & 26.1   & 59.4  & 69.9   & 25.6   & 59.7  & 65.4   & 41.3   & 54.5  & 70.8   & 44.2   & 58.5  \\ 
SEAM†\cite{wang2020self}       & VOC  & -  & - & 71.0   & 33.1   & 62.7  & 68.3   & 31.8   & 60.4  & 65.1   & 53.5   & 60.6  & 67.5   & 55.4   & 62.7  \\ 
SS† \cite{araslanov2020single}           & VOC   & - & - &  71.6   & 26.0   & 61.5  & 72.2   & 27.5   & 62.1  & 60.7   & 25.7   & 45.0  & 69.6   & 32.8   & 52.5  \\ 
EPS† \cite{lee2021railroad}          & VOC  & - &  - & 72.4   & 38.5   & 65.2  & 69.4   & 34.5   & 62.1  & 64.2   & 54.1   & 60.6  & 69.0   & 57.0   & 64.3  \\ 
WILSON†\cite{cermelli2022incremental}        & VOC  & - & - & 73.6   & 43.8   & 67.3  & 74.2   & 41.7   & 67.2  & 64.5   & \underline{54.3}   & 60.8  & 70.4   & 57.1   & 65.0  \\ 
RaSP\cite{roy2023rasp}  & VOC  & - &  -  & - &   -   &   -   &   76.2 & 47.0 & 70.0  &  -   &       -  &   -   &   72.3 & \underline{57.2} & \underline{65.9}  \\
FMWISS\cite{yu2023foundation} & VOC & VOC & 50 &  75.9   & \underline{50.8}   & 70.7  & \underline{78.4}   & \underline{54.5}   & \underline{73.3}  & \textbf{68.5}   & \textbf{58.2}   & \textbf{64.6}  & \textbf{73.8}   & \textbf{62.3}   & \textbf{69.1}  \\ 
Ours                   & VOC & \textbf{WEB} & 100 & \textbf{77.1}  & 49.1 &   \underline{71.1}    &    78.2    &  \textbf{54.9}  &   \underline{73.3}    &    \textbf{68.5}  &    52.0   &   \underline{61.7}      & 73.6 & 55.5 & 65.7 \\
Ours                   & VOC  & \textbf{WEB} &  500 &  \textbf{77.1} &    \textbf{53.0}     &    \textbf{72.0}     &  \textbf{78.8}     & {53.8}        &   \textbf{73.4}      &  67.1     &  52.1     &   {61.0}      & {73.0} & {55.2} & {65.3} \\ 

\midrule
WILSON \cite{cermelli2022incremental} & \textbf{WEB} & - & - & 75.2 & 45.5 & 68.9 & 74.2 & 43.8 & 67.8 & 66.9 & 47.3 & 58.6 & 72.0 & 49.7 & 62.1\\
Ours                  & \textbf{WEB} & - & - & 75.5  &  45.0   &   69.0    &  74.4  &  45.9   &  68.4   & \underline{68.3}      &  49.7       & 60.4        & \textbf{73.8} & 53.9 & 65.0 \\ 
Ours         & \textbf{WEB} & \textbf{WEB} & 100 & \underline{76.9}  &  47.0      &    70.5   &   78.3    &   47.8  &  71.7     &  68.0   &   50.0    &   60.4      & \underline{73.7} & 54.5 & 65.3 \\ 
Ours                   & \textbf{WEB} & \textbf{WEB} & 500 &  76.2  &    50.0    &   70.7    &    77.8   &  45.6   &   70.8    &  67.9     &  50.4      & 60.6        & 72.4 & 54.4 & 64.6 \\ 
\bottomrule
\end{tabular}
}
\label{tab:singlestep}
\end{table}

\subsubsection{Single-Step Multi-class Results}
The results for \textbf{15-5} and \textbf{10-10} on VOC (i.e., a single incremental step with 5 and 10 classes) are shown in Table \ref{tab:singlestep}.
For comparison purposes, we consider all other methods within the WILSS framework using the VOC dataset for training, while we compared our approach with WILSON \cite{cermelli2022incremental} also in the new setting, as it serves as the baseline framework.
From the table, several key points can be highlighted:
\begin{enumerate}
    \item Our rehearsal strategy, utilizing web data when training with the original dataset, outperforms all previous methods that do not store samples from the original dataset. In particular, in the 15-5 setting our approach outperforms all competitors and achieves an improvement of $5\%$ in the disjoint setup and of $6\%$ in the overlapped one w.r.t \cite{cermelli2022incremental}. Furthermore, unlike FMWISS \cite{yu2023foundation}, we did not rely on pixel-level supervision from pre-trained models, but solely on image-level supervision; On the 10-10 setup competitors are closer to our approach, however except \cite{yu2023foundation} that uses VOC supervision, we are the best in the disjoint setup and second best in the overlapped.
    \item In scenarios where no data from VOC is available, leveraging web data leads to comparable results to the state-of-the-art (up to $73.4\%$ in the 15-5 and $65.3\%$ in the 10-10 setup), with further improvements observed when utilizing rehearsal data (in this case we get very close to the case where VOC data is used, especially in the disjoint setting);
    \item Both of our strategies aimed at enhancing web data for old and new classes demonstrate improvements over the baseline \cite{cermelli2022incremental};
    \item We observed that employing a larger number of images (i.e., 500 as suggested in \cite{maracani2021recall}) leads to better results when training with data from the original distribution, but yields worse results when learning from the web.
\end{enumerate}

\begin{table}[t]
\centering
\begin{tabular}{cc}
\begin{minipage}[t]{0.45\textwidth}
    \centering
    \caption{Comparison on 60-20 setting on COCO-to-VOC. †: from \cite{cermelli2022incremental}.}
    \label{tab:coco}
    \centering
    \resizebox{\textwidth}{!}{
    \begin{tabular}{cccccccc|c}
    \toprule
    \multicolumn{1}{l}{} & \multicolumn{2}{c}{\textbf{Train}} & \multicolumn{2}{c}{\textbf{Memory}} &   \multicolumn{3}{c}{\textbf{COCO}} & \textbf{VOC} \\ \cmidrule{6-9} 
    \multicolumn{1}{c}{\textbf{Method}} & \textbf{1-60} &
    \textbf{61-80} &
    \textbf{Source} & \textbf{M} & 1-60 & 61-80 & All & 61-80 \\ 
    \midrule
    CAM† \cite{zhou2016learning} & COCO & VOC & - & - & 30.7 & 20.3 & 28.1 & 39.1 \\
    SEAM†\cite{wang2020self} & COCO & VOC & - & - & 31.2 & 28.2 & 30.5 & 48.0 \\
    SS† \cite{araslanov2020single} & COCO & VOC & - & - & 35.1 & 36.9 & 35.5 & 52.4 \\
    EPS† \cite{lee2021railroad} & COCO& VOC & - & - & 34.9 & 38.4 & 35.8 & 55.3 \\
    WILSON\cite{cermelli2022incremental} & COCO& VOC & - & - & 39.8 & \underline{41.0} & \underline{40.6} & 55.7 \\
    RaSP\cite{roy2023rasp} & COCO& VOC & - & - & \textbf{41.1} & \underline{41.0} & \textbf{41.6} & 54.4 \\
    FMWISS\cite{yu2023foundation}& COCO & VOC & COCO & 50 & 39.9 & \textbf{44.7} & \textbf{41.6} & \textbf{63.6} \\ 
    Ours& COCO & VOC & \textbf{WEB} & 100 & \underline{40.7} & 33.0 & 39.3 & \underline{56.7} \\
    \midrule
    WILSON \cite{cermelli2022incremental} & COCO & \textbf{WEB} & - & - & \underline{40.0} & \underline{33.2} & \underline{38.8} & \underline{48.9} \\
    Ours & COCO & \textbf{WEB} & - & - & \textbf{40.2} & \textbf{35.6} & \textbf{39.6} & \textbf{51.3} \\
    \midrule
    WILSON & COCO &\textbf{WEB} & \textbf{WEB}& 100 & \underline{40.6} & \underline{28.1} & \underline{38.0} & \underline{41.3}\\
    Ours & COCO & \textbf{WEB}& \textbf{WEB}& 100 & \textbf{40.9} & \textbf{35.8} & \textbf{40.2} & \textbf{49.5}\\
    \bottomrule
    \end{tabular}
     }
\label{tab}
\end{minipage}
&
\begin{minipage}[t]{0.55\textwidth}
\centering
    \caption{Comparison on single-step multi-class settings on PASCAL-VOC.}
    \resizebox{\textwidth}{!}{
    \begin{tabular}{ccccccc|ccc}
    \toprule
     &  &  &  & \multicolumn{3}{c}{15-1} & \multicolumn{3}{c}{10-1} \\ \cmidrule{5-10}
    \textbf{Method} & \textbf{Train} & \textbf{Memory} & \textbf{M} & 1-15 & 16-20 & All & 1-10 & 11-20 & All \\
    \midrule
    WILSON \cite{cermelli2022incremental} & VOC & - & - & 0.0 & 2.3 & 0.6 & 0.0 & 0.2 & 0.1\\
    RaSP\cite{roy2023rasp} & VOC & - & - & 17.7 & 0.9 & 13.2 & 2.0 & 0.7 & 1.3 \\
    RaSP \cite{roy2023rasp} & VOC & VOC & 100 & 63.3 & 28.3 & 56.0 & 38.9 & 30.9 & 36.9 \\
    WILSON \cite{cermelli2022incremental} & VOC & ImageNet & 100 & \textbf{75.7} & 32.9 & 65.9 & \textbf{66.8} & 34.9 & \underline{52.3} \\
    RaSP \cite{roy2023rasp} & VOC & ImageNet & 100 & \textbf{75.7} & 35.2 & 66.6 & \textbf{66.8} & 39.1 & \textbf{54.4} \\
    \midrule
    WILSON & VOC & WEB & 100 & 73.9 & 29.9 & 63.6 & 63.0 & 35.6 & 49.9 \\
    Ours & VOC & WEB & 100 & 73.6 & 36.2 & 65.1 & \underline{63.9} & \underline{39.2} & 52.0\\ 
    \midrule
    WILSON & WEB & WEB & 100 & 74.4 & 41.6 & \underline{67.2} & 61.0 & 28.3 & 42.9 \\
    Ours & WEB & WEB & 100 & \underline{74.5} & \textbf{41.9} & \textbf{67.4} & 63.7 & \textbf{42.4} & \textbf{54.4} \\ 
    \bottomrule
    \end{tabular}
    \label{tab:multistep}
    }
\end{minipage}
\end{tabular}
\end{table}

Results on the COCO-to-VOC benchmark are shown in \cref{tab:coco}. Our replay strategy achieves a mIoU of 56.7\% on VOC, outperforming exemplar-free competitors, second only to \cite{yu2023foundation} which however stores COCO samples.
We demonstrate learning from web data as a viable alternative to controlled data collection, outperforming models trained on original datasets. In settings using web data for the incremental step, we achieve accuracies on VOC around 50\% without exposure to VOC data distribution. Incorporating web data for both old and new classes increases COCO mIoU, though this improvement does not fully transfer to VOC. However, results surpass WILSON trained with the same data on both datasets. Please note that our method focuses on obtaining web data matching the original distribution, i.e., aligned with COCO, for both old and new classes.
Future investigations could explore optimal strategies to balance between matching the original distribution and generalization to unseen domains. Nonetheless, new web images allow incremental training, while old web images remain advantageous, particularly in multi-step single-class learning scenarios, where competitors necessitate the incorporation of old classes' samples.

\subsubsection{Multi-Step Single-class Results}
Results for the \textbf{15-1} and \textbf{10-1} overlapped settings are in Table \ref{tab:multistep}. Notably, among competitors, only \cite{roy2023rasp} addresses these settings. However, it is noteworthy that \cite{roy2023rasp} still relied on handcrafted datasets (i.e., VOC or ImageNet), rather than unsupervised data.
We can observe that: 
\begin{enumerate}
    \item Leveraging web data as rehearsal enables WILSON to learn within the multi-step single-class scenario;
    \item Our approach outperforms the standard strategy based on \cite{cermelli2022incremental} of around $1.5\%$ in the 15-1 and $2\%$ in the 10-10 setup;
    \item Although remembering past classes from human polished and large curated datasets is still the best option, our method manages to narrow the performance gap getting very close to competitors exploiting ImageNet  rehearsal;
    \item  The most notable advantage of our approach is demonstrated in the final setting of \Cref{tab:multistep}, where learning new knowledge from the web achieves superior performance compared to training from the original dataset and allows us to outperform competitors in the 15-1 setting and to get exactly the same performances of \cite{roy2023rasp} in the 10-1 but using web data instead of VOC and ImageNet for the competitor. This is very relevant as some classes in VOC are short in data, suggesting the potential viability of our setup. 
\end{enumerate}

\begin{figure}[tp]
\centering
\vspace{-0.2cm}
\includegraphics[trim=2.5cm 5cm 2.5cm 3cm, clip, width=0.95\linewidth]{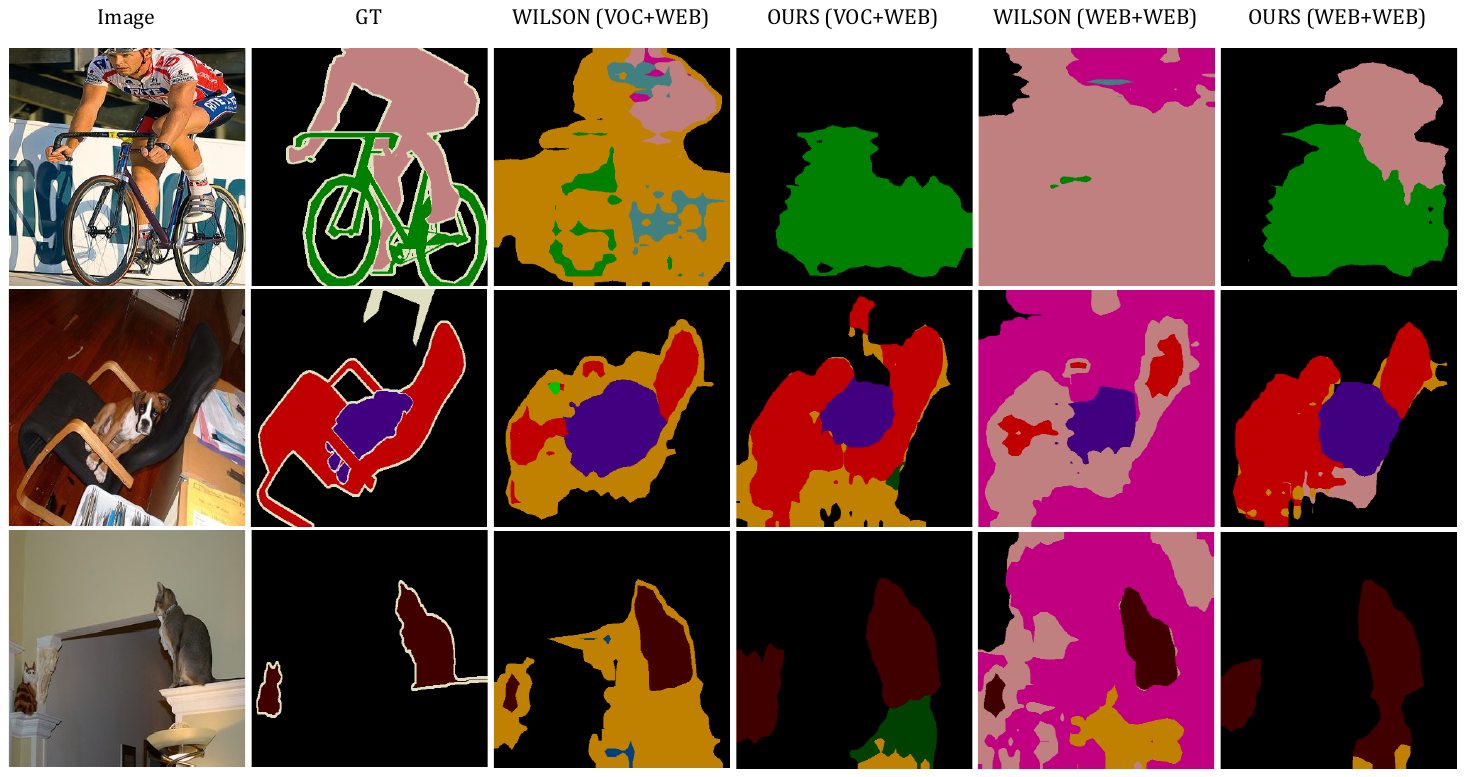}
\vspace{-0.2cm}
\caption{Qualitative results in the overlapped 10-1 setting.
} 
\label{fig:qualitative1}
\end{figure}

Figure \ref{fig:qualitative1} shows some visual examples for the most challenging 10-1 setting. Notice how they confirm the numerical evaluation with our approach being the only one able to properly recognize the main objects in the scene while WILSON show a large number of wrong detections. The shape of the objects is good even if not extremely precise  - yet recall that there is no segmentation ground truth in the incremental steps - and in any case much better than the main competitor.

\textbf{Ablation Results} In order to analyze the impact of our web rehearsal strategies we performed some ablation studies. 
Table \ref{tab:ablationWeb} shows the impact of the two newly introduced modules for the web learning strategy proposed in this paper. The domain discriminator in the Fourier domain allows for a gain of around $1\%$ in the 10-10 setup, while the new caption labeling strategy for the localizer allows for a bit larger improvement of $1.5\%$ in the same setup. Notice how the two strategies are complementary and their combined usage leads to an improvement of around $3\%$. 

\begin{table}[t]
\centering
\begin{tabular}{cc}
\begin{minipage}{0.5\textwidth}
    \centering
    \caption{Ablation study on the web learning strategies: Pascal-VOC Overlapped.}
    \label{tab:ablationWeb}
    \resizebox{\textwidth}{!}{
    \begin{tabular}{cccccc|ccc}
    \toprule
    \multicolumn{3}{c}{\textbf{Learning from Web}} & \multicolumn{3}{c}{10-10} & \multicolumn{3}{c}{15-5} \\ \cmidrule{4-9}
    Domain filtering & Caption Labeling & & 1-10 & 11-20 & All & 1-15 & 16-20 & All \\
    \midrule
     &  &  & 72.0 & 49.7 & 62.1 & 74.2 & 43.8 & 67.8 \\
    \checkmark &  &  & 72.0 & 51.9 & 63.1 & 73.7 & \underline{46.3} & 68.0 \\
     & \checkmark & & \underline{72.6} & \underline{52.1} & \underline{63.6} & \underline{73.8} & \textbf{46.7} & \underline{68.2} \\
    \checkmark & \checkmark & & \textbf{73.8} & \textbf{53.9} & \textbf{65.0} & \textbf{74.4} & 45.9 & \textbf{68.4} \\
    \bottomrule
    \end{tabular}
    }
\label{tab
}
\end{minipage}
&
\begin{minipage}{0.5\textwidth}
\centering
    \caption{Ablation study on the rehearsal strategies on Pascal-VOC Overlapped.}
    \label{tab:ablationR}
    \resizebox{\textwidth}{!}{
    \begin{tabular}{cccccc|ccc}
    \toprule
    \multicolumn{3}{c}{\textbf{Rehearsal Strategy}} & \multicolumn{3}{c}{10-10} & \multicolumn{3}{c}{15-5} \\ \cmidrule{4-9}
    Caption query & Caption Filtering &  & 1-10 & 11-20 & All & 1-15 & 16-20 & All \\
    \midrule
     &  &  & 71.6 & 53.1 & 63.6 & 76.0 & 47.4 & 69.8 \\
    \checkmark &  &  & \underline{72.2} & \underline{55.1} & \underline{64.9} & 77.6 & \textbf{53.9} & \underline{72.6} \\
     & \checkmark & & 71.5 & 52.9 & 63.5 & \textbf{78.8} & {47.8} & {71.9} \\
    \checkmark & \checkmark & &  \textbf{73.0} & \textbf{55.2} & \textbf{65.3} & \underline{78.7} & \underline{53.7} & \textbf{73.4} \\ 
    \bottomrule
    \end{tabular}
    }
\end{minipage}
\end{tabular}
\end{table}

Table \ref{tab:ablationR} shows instead the impact of the caption querying and caption filtering proposed in this paper. 
Using just class names as queries we achieve a mIoU of $63.6\%$ in the overlapped 10-10 setting and $69.8\%$ in the 15-5. Using the captions for web querying allows for improved performances in {both 10-10 and 15-5 setting.}
Generating new captions to filter the images leads to an improvement in the 15-5 setup while the 10-10 performances remain roughly the same. Note that, when no queried caption is available, the filter strategy solely relies on the class name, reducing the strength of the text description. Finally, the potential for gathering more reliable data is achieved when combining the two strategies, leading to an improvement of $1.7\%$ in the 10-10 scenario and $3.6\%$ in the 15-5 scenario.
\section{Limitations}
\label{sec:limit}
The present framework relies on supervising through a foundation model, similar to other setups employing such models. While it is reasonable to assume that a large model encompasses knowledge of concepts connected to future classes one aims to learn, this assumption may not hold. Subsequent research could address this concern. 
A practical implementation of the system should download web data on the fly while the learning goes on, how to optimize the bandwidth usage and the download time requirements is also left as future research.

\section{Conclusions}
\label{sec:concl}
This paper introduces a novel approach to address incremental learning in semantic segmentation by leveraging web data instead of manually selected and annotated datasets. We examine both the existing WILSS setup from literature and a novel approach where supervised data is utilized only in the initial step, followed by learning driven solely by language cues and web images. Our method utilizes a VLM to extract meaningful captions for data samples, which are then used to guide web querying and select suitable web images. By integrating this with state-of-the-art continual learning strategies, we achieve remarkable performance, demonstrating the effectiveness of web data in replacing manually curated datasets in continual learning tasks. Future research will address the pointed out limitations and consider the extension of the approach to other segmentation architectures or to tasks behind semantic segmentation.

\vspace{0.2cm}

\textbf{Acknowledgements}: This work was partially supported by the National Natural Science Foundation of China (62371356), the European Union, Italian National Recovery, Resilience Plan of NextGenerationEU, partnership ``Telecommunications of the Future'' (PE00000001-  ``RESTART''), Fundamental Research Funds for the Central Universities of China (ZYTS24005) and the National Defense Basic Scientific Research Program of China (JCKY2021413B005).

\bibliographystyle{splncs04}
\bibliography{main}







\title{Learning from the Web: Language Drives Weakly-Supervised Incremental Learning for Semantic Segmentation\\  \textit{Supplementary Material}}

\titlerunning{Learning from the Web: Language Drives Weakly-Supervised ILSS}

\author{Chang Liu\inst{1}\orcidlink{0000-0002-5321-2264} \and
Giulia Rizzoli\inst{2}\orcidlink{0000-0002-1390-8419} \and
Pietro Zanuttigh\inst{2}\orcidlink{0000-0002-9502-2389} \and
Fu Li\inst{1}\orcidlink{0000-0003-0319-0308} \and
Yi Niu\inst{1}\orcidlink{0000-0002-7359-276X}\thanks{Corresponding author}}

\authorrunning{C.~Liu, G.~Rizzoli et al.}

\institute{School of Artificial Intelligence, Xidian University, China \and
Department of Information Engineering, University of Padova, Italy
}

\maketitle

This document contains the supplementary material for the paper \textit{Learning from the Web: Language Drives Weakly-Supervised Incremental Learning for Semantic Segmentation}.
Firstly, we provide a detailed description of the knowledge distillation losses of the localizer and segmentation model.
Then, to present further validation, we showcase 
 1) an upper bound of the proposed method that directly uses PASCAL-VOC as the replay source and 2) the per-step results on the VOC dataset obtained in the most challenging setting (10-1), which involves the highest number of incremental steps.
Besides, we also explore the robustness of the proposed method to the choice of different VLMs.
Finally, we motivate our framework by showing ablation studies, quantitative and qualitative support for each component of the method, including the Fourier discriminator, caption pseudo-labeling, caption downloading, and caption filtering.

\section{Knowledge Distillation Losses}
The localizer $L$, introduced in \cite{cermelli2022incremental}, is used to provide a pseudo-supervision for the main segmentation model. It shares the same encoder $E$ with main model and predicts a score $\textbf{y}_L$ for all the classes $|\mathcal{Y}|$:
\begin{equation}
 \textbf{y}_{L} = (E \circ L) \in \mathbb{R}^{|\mathcal{I}| \times |\mathcal{Y}|}
\end{equation}

The training objective of the localizer has a dual role: 1) learning new classes from image-level labels and 2) refining old classes using the previous segmentation model's output.
The former follows the classification task, in which the class score is normally calculated with  global average pooling (GAP). However, using GAP tends to encourage all the pixels to identify with the target classes, which weakens diversity and is thus not suitable for the segmentation task. Based on the above concern, Araslanov et al. \cite{araslanov2020single} proposed Global Weighted Pooling (GWP) and focal penalty to calculate a more precise classification score.
While effective for new classes, this approach still struggles with localizing previous classes.  
To this extent, WILSON \cite{cermelli2022incremental} introduced a knowledge distillation-based localization prior loss ($KDL$) that leverages the previous segmentation model's output as pseudo-supervision for old classes:
\begin{equation}
    \begin{split}
    \mathcal{L}_{KDL}(\textbf{z},\textbf{y}_D^{(t-1)})&= -\frac{1}{|\mathcal{Y}^{t-1}||\mathcal{I}|}\sum_{i\in\mathcal{I}}\sum_{c\in\mathcal{Y}^{t-1}}y_{(i) D}^{c(t-1)}log(\sigma(z_i^c))+ \\
    &+ (1-y_{(i) D}^{c(t-1)})log(1-\sigma(z_i^c))
    \end{split}
\end{equation}

where $\sigma$ denotes the logistic function applied to $\textbf{z}$, which represents $\textbf{y}_L$ prior to the softmax operation, and $\textbf{y}_D^{(t-1)}$ is the output of the segmentation model from step $t-1$.
The $KDL$ loss provides pixel-wise supervision to the localizer, enhancing the precision of the localizer's class activation map. 
Similarly, a distillation loss $\mathcal{L}_{KDE}$ is applied to the segmentation model: it   computes the mean-squared error between the features extracted by the encoder of the current step $E^t$ and the one of the previous step  $E^{t-1}$:

\begin{equation}
    \mathcal{L}_{KDE} = \frac{1}{|\mathcal{I}|} \sum_{i \in \mathcal{I}} \|e_i^t - e_i^{t-1}\|
\label{eq:kde_loss}
\end{equation}
where $e$ is the feature vector from the encoder $E$.

\section{Additional Results}

\subsection{VOC as Rehearsal data}
{We present in Table \ref{tab:voc} the results obtained by utilizing VOC directly for replay, which serves as an upper bound for our approach. Note how our approach closely approaches the bound, particularly in the 15-5 setting, demonstrating the effectiveness of utilizing WEB data. 
Furthermore, when employing 50 VOC images — equivalent to the number used by the FMWISS \cite{yu2023foundation} competitor — our approach yields slightly better or similar results. Specifically, we observe a slight decrease of 0.3 mIoU points in the 10-10 setting and an improvement of 0.8 in the 15-5 setting compared to FMWISS.
}

\begin{table}[h]
\centering
\caption{Results using VOC as replay source on PASCAL-VOC overlapped setup.}
\label{tab:voc}
\begin{tabular}{cccc|ccc|ccc} \toprule
& \multicolumn{3}{c|}{}&  \multicolumn{3}{c}{\textbf{10-10}} &  \multicolumn{3}{c}{\textbf{15-5}} \\ \cmidrule{5-10}
\textbf{Method} & \textbf{Train} & \textbf{Memory} & \textbf{M} & 1-10 & 11-20 & all & 1-15 & 16-20 & all\\\midrule
FMWISS\cite{yu2023foundation} & VOC & VOC & 50 & \textbf{73.8} & 62.3 & \underline{69.1} & \textbf{78.4} & 54.5 & 73.3 \\ \midrule
Ours & VOC & WEB & 100  & \underline{73.6} & 55.5 & 65.7 & \underline{78.2} & 54.9 & 73.3 \\ 
Ours & VOC & VOC & 50 & 67.4 & \textbf{68.1} & {68.8} & 76.5 & \underline{63.4} & \underline{74.1} \\ 
Ours & VOC & VOC & 100 & 71.9 & \underline{66.2} & \textbf{70.1} & 76.5 & \textbf{65.5} & \textbf{74.6} \\ \bottomrule 
\end{tabular}
\end{table}

\subsection{Class-wise Per-step mIoU}
\begin{figure}[htpb]
\centering
\vspace{-0.2cm}
\includegraphics[trim=17cm 11cm 21cm 10cm, clip, width=0.9\linewidth]{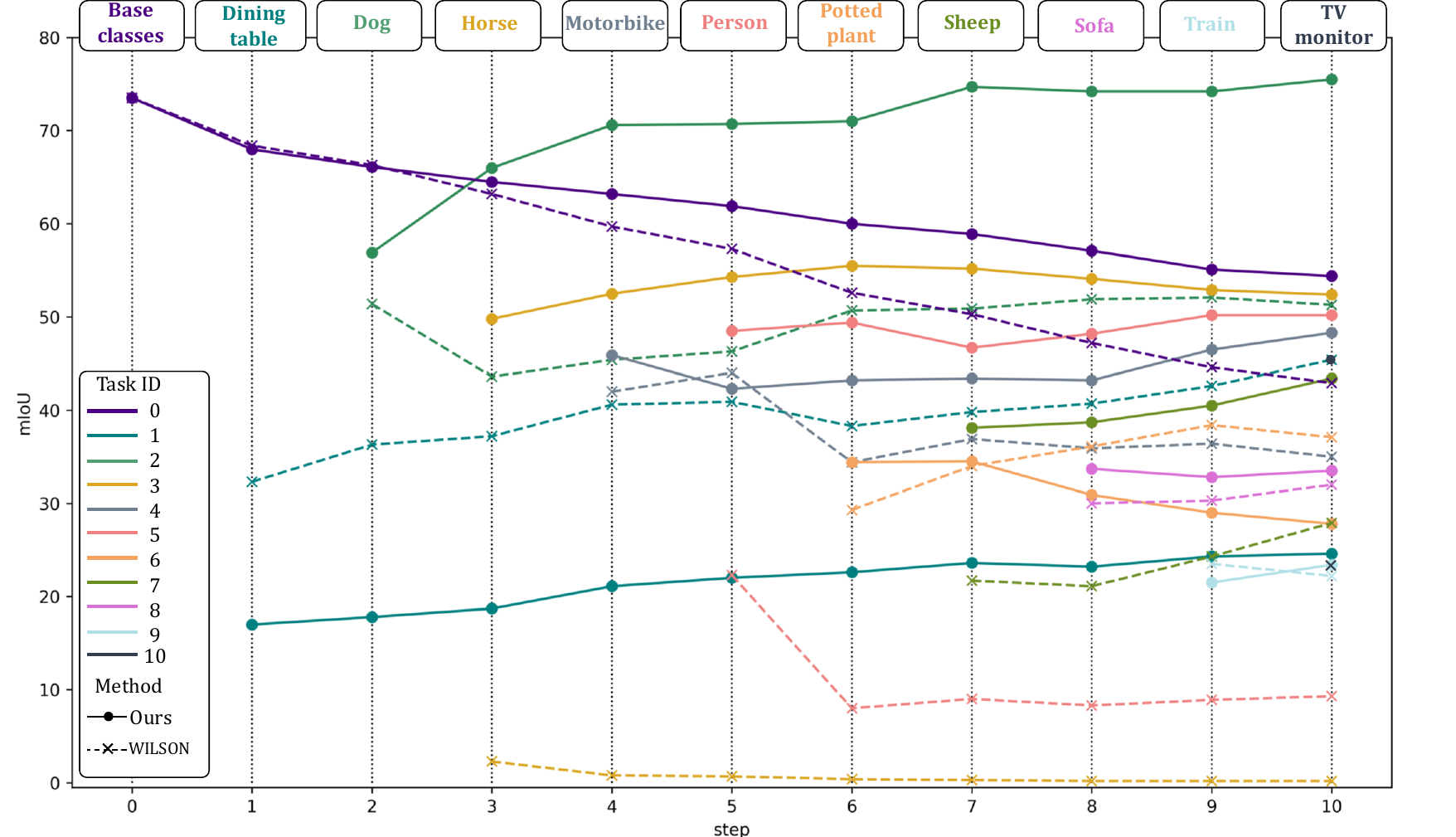}
\vspace{-0.2cm}
\caption{Per-task and per-step mIoU for the 10-1 VOC multi-step \textit{overlap} incremental setting (WEB+WEB).} 
\label{fig:per-step}
\vspace{-0.35cm}
\end{figure}

\cref{fig:per-step} illustrates the per-step performance of our approach (solid lines) and compares them with WILSON \cite{cermelli2022incremental} (dashed lines), highlighting the consistent improvement of our method over WILSON across the various steps of the learning sequence. We outperform the competitor by a large margin in most classes (including the initial set of classes learned in the first step) even if a very few challenging ones persist, such as dining table and potted plant.

\subsection{Evaluation with a different VLM}
While the results in the paper were obtained using the OpenFlamingo \cite{awadalla2023openflamingo} model, our approach demonstrates versatility to different Vision-Language Models (VLMs).  In \cref{tab:vlm} we showcase the results of replacing OpenFlamingo with BLIP \cite{li2022blip}. The performances are slightly better in the 10-10 setting and slightly worse on average in 15-5. Notably, these performance differentials fall within a narrow range of [0.2, 0.5] mIoU points.
These minimal variations confirm that our proposed method is not overfitted on a particular VLM configuration. Rather, it shows the robustness and generalizability in combination with different models.

\begin{table}[h]
\centering
\caption{Comparison with BLIP VLM in the PASCAL-VOC overlapped setup.}
\label{tab:vlm}
\begin{tabular}{cccc|ccc|ccc} \toprule
  & \multicolumn{3}{c}{ }  & \multicolumn{3}{c}{\textbf{10-10}} & \multicolumn{3}{c}{\textbf{15-5}} \\ 
\textbf{Method} & \textbf{Train} & \textbf{Memory} & \textbf{M} & 1-10 & 10-20 & all & 1-15 & 16-20 & all \\ \midrule
OpenFlamingo & VOC & WEB & 100 & \textbf{73.6} & 55.5 & 65.7 & \textbf{78.2} & \textbf{54.9} & \textbf{73.3} \\ 
BLIP & VOC & WEB & 100 & 73.3 & \textbf{56.3} & \textbf{65.9} & 78.1 & 54.1 & 73.0 \\  
OpenFlamingo & WEB & - & - & \textbf{73.8} & 53.9 & 65.0 & \textbf{74.4} & 45.9 & \textbf{68.4} \\ 
BLIP & WEB & - & - & 72.2 & \textbf{56.1} & \textbf{65.3} & 73.5 & \textbf{46.4} & 67.9 \\
OpenFlamingo & WEB & WEB & 100 & \textbf{73.7} & 54.5 & 65.3 & \textbf{78.3} & 47.8 & {71.7} \\ 
BLIP & WEB & WEB & 100 & 73.0 & \textbf{55.8} & \textbf{65.6} & 77.1 & \textbf{51.9} & \textbf{71.8} \\
\bottomrule
\end{tabular}
\end{table}

\newpage

\section{Framework motivation}
In this section, we aim to support each component of our method by presenting some additional quantitative and qualitative results.

\subsection{Learning new Knowledge from Web}

\subsubsection{Fourier Domain-based Discriminator}
To extract images with statistics resembling the original ones, we have trained a discriminator network, helping us to select web images for the new classes that closely resemble those from the original dataset. We trained the discriminator until it reached 80\% accuracy on VOC data to ensure a proper behaviour of this model.
To validate the efficacy of training the discriminator in the Fourier domain, we compare in \cref{tab:abl_fft-vs-pix} the performance of our approach with the baseline strategy in the pixel domain, i.e., substituting Eq. 6 of the main paper with $(p_{ds}, \, p_{web}) = M_D(\mathbf{x})$.
In the frequency domain, the general style information of the image is primarily contained in lower frequencies, while specific content is typically associated with higher frequencies \cite{yang2020fda}. This separation makes distinguishing dataset-wide properties from image-specific content easier than using approaches in the pixel domain or exploiting the Fourier phase  (see \cref{tab:abl_fft-vs-pix}).
Using the amplitude values of the Fourier Transform in place of the original sample, values lead to a mIoU improvement in learning new classes of $3.3\%$ on 10-10 and $2.3\%$ on 15-5 and of $1.2\%$ and $0.9\%$ in the average value.
Moreover, this strategy demonstrates good generalization, with accuracy on new (unseen) classes during incremental steps reaching $76-77\%$, quite close to the $80\%$ achieved on training classes. In contrast, with a pixel-domain discriminator we only achieved $64-66\%$ accuracy.

\begin{table}[htbp]
    \centering
    \caption{Sample selection strategy in the spatial and Fourier domain: mIoU in the PASCAL-VOC overlapped setup.}
    \begin{tabular}{cccc|ccc}
        \toprule
          & &  \textbf{10-10} & & & \textbf{15-5} \\ \cmidrule{2-7}
          \textbf{Domain} & 1-10 & 11-20 & all & 1-15 & 16-20 & all \\ \midrule
          Pixel & \textbf{72.6} & 48.6 & 61.9 & 73.2 & 44.0 & 67.1 \\
          \midrule
          Fourier-Phase & \underline{72.4} & \underline{50.1} & \underline{62.4} & \textbf{73.9} & 43.8 & \underline{67.6}\\
         Fourier-Ampl. & 72.0 & \textbf{51.9} & \textbf{63.1} & \underline{73.7} & \textbf{46.3} & \textbf{68.0} \\
         \bottomrule
    \end{tabular}
    \label{tab:abl_fft-vs-pix}
\end{table}

\subsubsection{Caption Labeling}
\begin{figure}[htbp]
\centering
\vspace{-0.2cm}
\includegraphics[trim=3cm 11.5cm 3.5cm 3cm, clip, width=\linewidth]{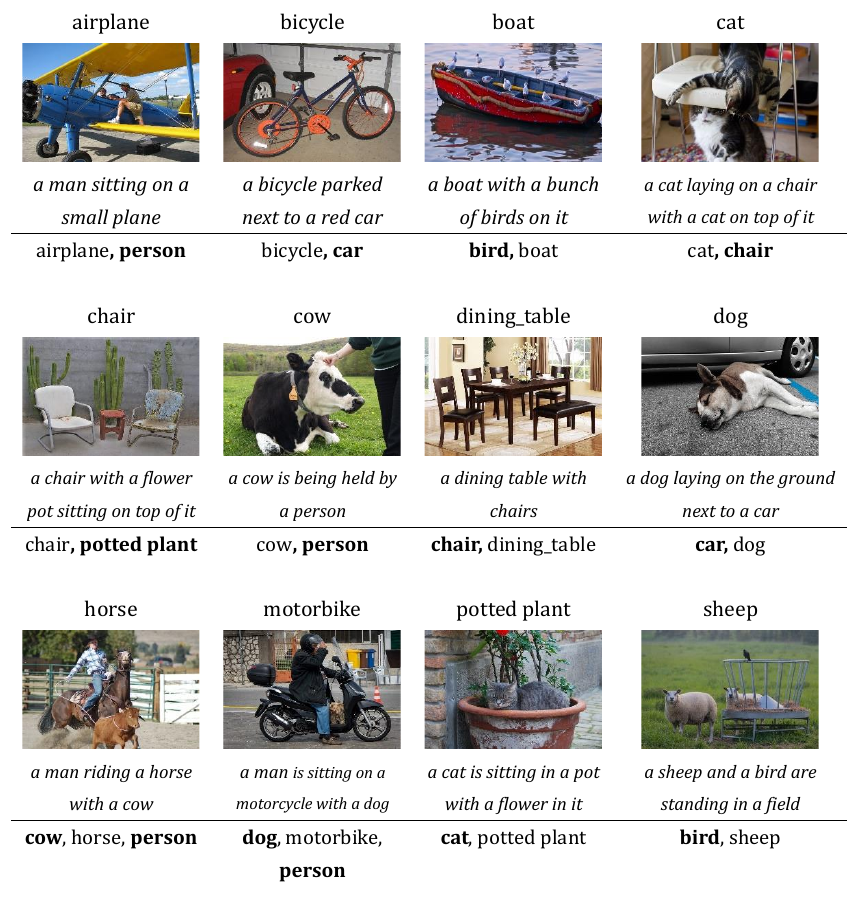}
\vspace{-0.2cm}
\caption{Image-level labels generated from captions for COCO-to-VOC incremental step classes. For each sample we show (from top to bottom) the queried class name, a thumbnail of the image, the generated caption and the final image-level label.} 
\label{fig:cap-labeling}
\vspace{-0.4cm}
\end{figure}

Although the images for the new classes are queried by the class name (Eq. 4), the downloaded image might also contain classes that have already been learned or are currently being learned. The latter case is particularly evident in the COCO-to-VOC scenario, where 20 classes are being learned simultaneously. In \cref{fig:cap-labeling} we provide examples demonstrating how, by analyzing captions and cross-referencing words with a reference dictionary (see \cref{tab:synonyms}), we can derive multi-class labels.

\begin{table}[]
    \centering
    \caption{Set of words $\mathcal{W}$ corresponding to each class $c$ in PASCAL-VOC.}
    \begin{tabular}{|c|c|c|c|c|} \cline{1-2} \cline{4-5}
        class & synonyms && class & synonyms \\ \cline{1-2} \cline{4-5}
          airplane & plane, jetliner && dining table & - \\ 
          bicycle & bike && dog & - \\
          bird & \makecell[c]{parrot, duck, flamingo, \\ swan, seagull, chicken} && horse & - \\
          boat & ship && motorbike & motorcycle \\ 
          bottle & - && person & \makecell[c]{man, men, woman, \\ women, people, baby}   \\
          bus & - && potted plant & pot of plant, pot of flower\\
          car & - && sheep & -\\
          cat & - && sofa & couch \\
          chair & - && train & train car\\
          cow & - && tv & television, monitor, tv monitor\\
          \cline{1-2} \cline{4-5}
    \end{tabular}
    \label{tab:synonyms}
\end{table}

\subsection{Reharsal strategies}

\subsubsection{Caption-based Querying.}
\begin{figure}[htpb]
\centering
\vspace{-0.2cm}
\includegraphics[trim=3cm 2.9cm 3.5cm 2.5cm, clip, width=0.8\linewidth]{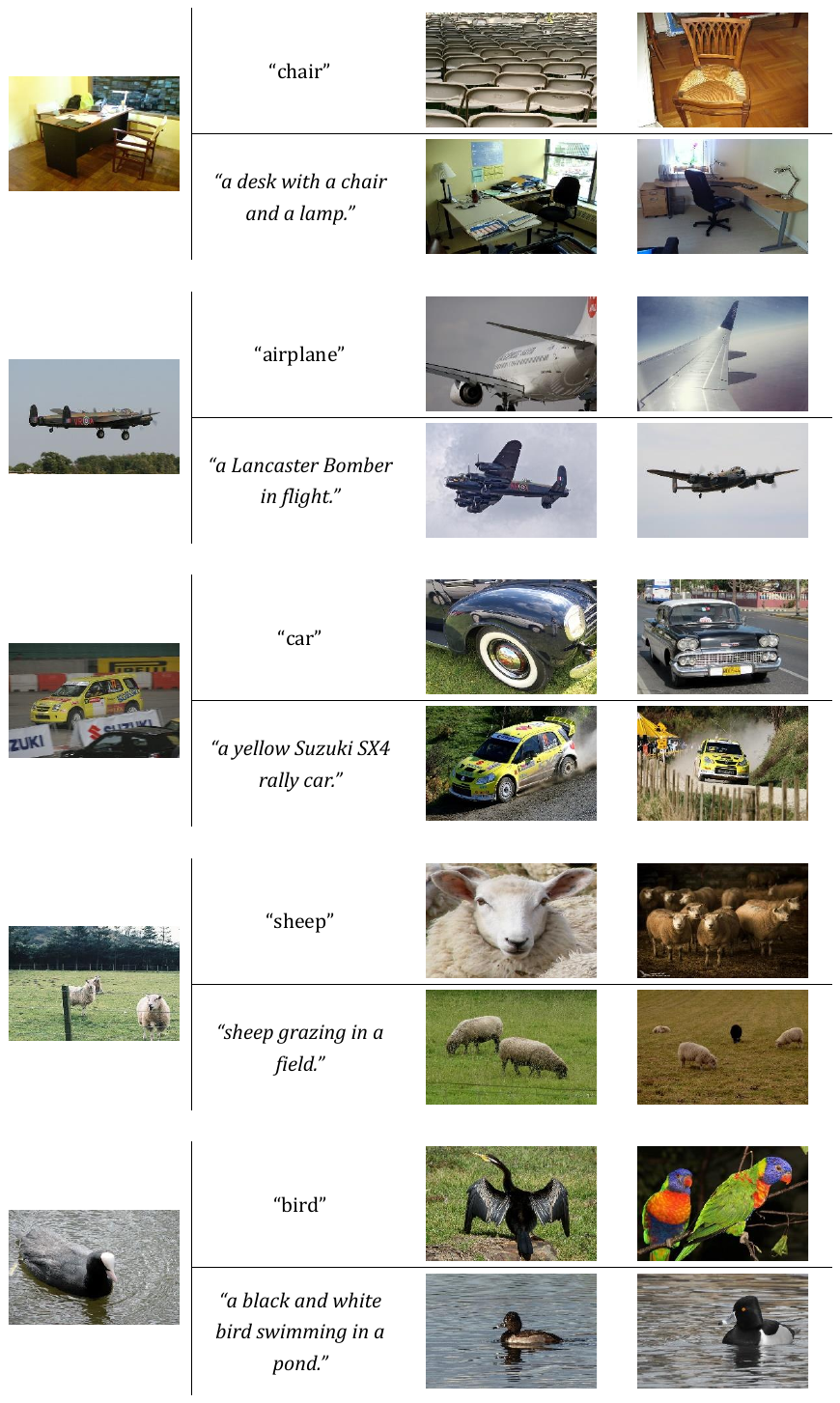}
\vspace{-0.2cm}
\caption{Comparison between caption and class-based web query strategy. (Left) Pascal sample and its related class and caption. (Upper right) Web samples downloaded with class name. (Lower right) Web samples downloaded with caption.} 
\label{fig:cap-download}
\vspace{-0.35cm}
\end{figure}

In \cref{fig:cap-download} we show some examples of downloaded images obtained through class-name  (i.e., Eq. 4) and through the caption (i.e.,  Eq. 8). Comparing them visually reveals that the caption method retrieves images more similar to the PASCAL-VOC ones without needing the storage of sensitive data. 
Specifically, in the first example the caption preserves related objects from the original scene (i.e., the lamp and the desk). In the subsequent two instances, it maintains the specific plane and car model.
Finally, in the last two cases, it keeps the original background (i.e., the grass and the water).
Notably, the background class - including not only the \textit{stuff} classes but all the objects that are not part of the training - plays an important role in causing the distribution shift as underlined in previous CILSS research works \cite{cermelli2020modeling}.

\subsubsection{Caption-based Filtering.}
Although our web retrieval scheme is efficient, there is no definite assurance that the downloaded images actually contain the objects referenced in the respective captions $q'$. This discrepancy can be seen in \cref{fig:cap-filter}, where the leftmost images show the original PASCAL-VOC image for which the captions were generated, and each upper row contains examples of downloaded images for which the re-generated captions $q''$ do not match the original one. 
Our method verifies this consistency by not only retaining the main objects (such as the dog, the sofa, and the person) but also capturing contextual details in some instances, for instance, the living room and the mountain. More examples are shown in \cref{fig:cap-all-replay} that compares PASCAL-VOC samples with similar ones retrieved from the web. {Additionally, the ablation on parameter $T$ of the caption-filtering is in \cref{tab:T}. Results are stable for $T$ in  $[0.5,0.7]$, with $T\!=\!0.6$ leading to the best tradeoff between old and new classes. 
{Furthermore, we conducted an ablation study on the number of selected words from the caption at step 2 of the filtering strategy. Our findings indicate that considering only 2 words proved to be slightly preferable. This outcome may be attributed to the characteristics of the VOC dataset, where less than 1/10th of the images contain more than two classes. The parameter can be tuned depending on the target distribution data. Nevertheless, when the distribution is unknown, all the nouns can be kept, leading to similar results as shown in \cref{tab:N}.
}

\begin{table}[h]
\centering
\caption{Ablation on threshold $T$ in caption-based filtering: mIoU in the
PASCAL-VOC overlapped setup.}
\label{tab:T}
\begin{tabular}{c|ccc|ccc} \toprule
\textbf{} &  \multicolumn{3}{c}{\textbf{10-10}} &  \multicolumn{3}{c}{\textbf{15-5}} \\ \cmidrule{2-7}
\textbf{T} & 1-10 & 10-20 & all & 1-15 & 15-20 & all \\ \midrule
0.0 & 72.2 & 55.1 & 64.9 & 77.6 & \underline{53.9} & 72.6\\ 
\midrule
0.5 & 72.5 & \textbf{55.8} & 65.3 & 77.8 & \underline{53.9} & 72.8 \\ 
Ours (0.6) & \textbf{73.6} & \underline{55.5} & \textbf{65.7} & \textbf{78.2} & \textbf{54.9} & \textbf{73.3} \\ 
0.7 & \underline{73.4} & \underline{55.5} & \underline{65.6} & \underline{78.1} & 53.6 & \underline{72.9} \\
\bottomrule
\end{tabular}
\end{table}

\begin{table}[]
    \centering
    \caption{Ablation on the number of nouns $N$ in caption-based filtering: mIoU in the PASCAL-VOC overlapped setup.}
    \label{tab:N}
    \begin{tabular}{c|ccc|ccc}
    \toprule
     \textbf{} & \multicolumn{3}{c|}{\textbf{10-10}} & \multicolumn{3}{c}{\textbf{15-5}} \\ \cmidrule{2-7}
    \textit{N} & 1-10 & 10-20 & all & 1-15 & 15-20 & all \\ \midrule
    1 & 73.0 & 55.8 & 65.6 & 77.9 & 52.9 & 72.6 \\
    Ours (2) & \textbf{73.6} & 55.5 & \underline{65.7} & \textbf{78.2} & \textbf{54.9} & \textbf{73.3} \\
    3 & \underline{73.1} & \underline{55.9} & \underline{65.7} & \underline{78.1} & 53.8 & \underline{73.0}  \\
    ALL & 72.5 & \textbf{56.8} & \textbf{65.8} & 77.8 & \underline{54.0} & 72.8 \\ \bottomrule
    \end{tabular}
\end{table}

\begin{figure}[htpb]
\centering
\vspace{-0.2cm}
\includegraphics[trim=2cm 3cm 2cm 2cm, clip, width=0.9\linewidth]{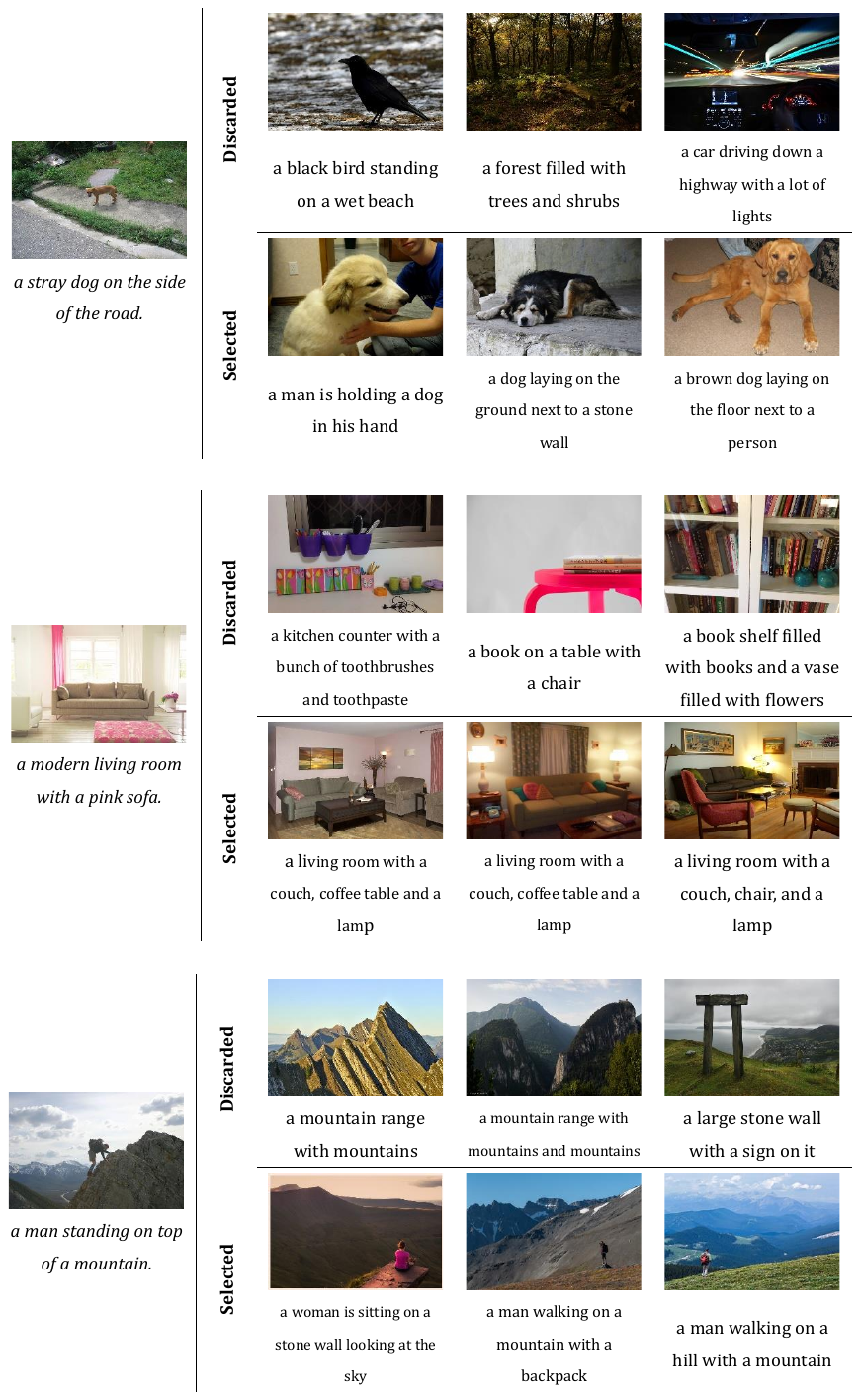}
\vspace{-0.2cm}
\caption{Examples of selected and discarded images by regenerating caption for the downloaded web samples. (Left) Dataset image and its caption; (Upper right) Discarded samples and corresponding captions. (Lower right) Selected samples and corresponding captions.} 
\label{fig:cap-filter}
\vspace{-0.35cm}
\end{figure} 

\begin{figure}[htpb]
\centering
\vspace{-0.2cm}
\includegraphics[trim=2cm 4cm 2cm 2cm, clip, width=\linewidth]{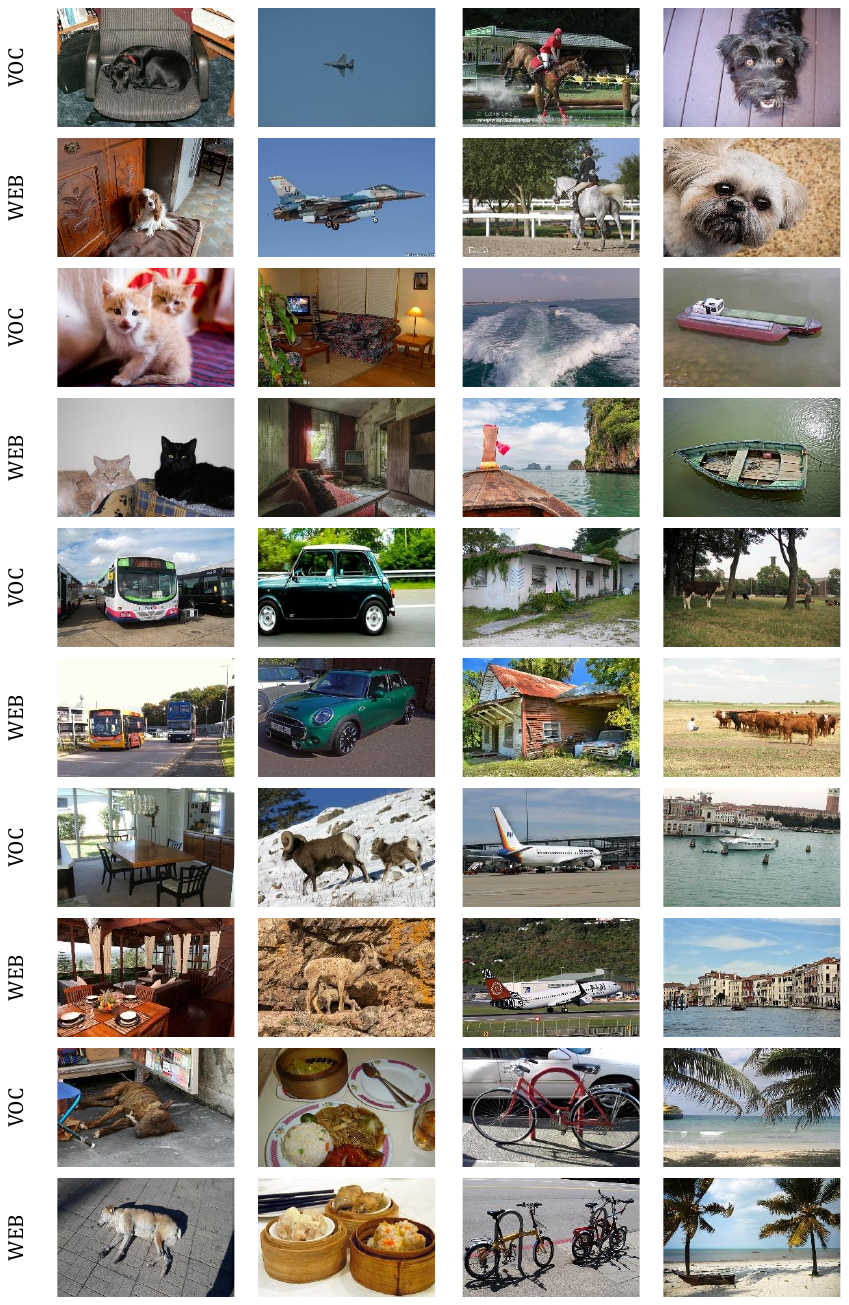}
\vspace{-0.2cm}
\caption{Replay samples selected with caption model compared with PASCAL-VOC ones.} 
\label{fig:cap-all-replay}
\vspace{-0.35cm}
\end{figure}

\clearpage


\end{document}